\documentclass[11pt,letterpaper]{article}
\usepackage{cogsys}
\usepackage[T1]{fontenc}
\usepackage{times}
\usepackage[pdftex]{graphicx} 
\usepackage[normalem]{ulem}

\usepackage{natbib}
\usepackage{todonotes} 
\newcommand{\dustin}[1]{\todo[inline,color=green!20!white]{\textbf{Dustin:} #1}}
\newcommand{\adam}[1]{\todo[inline,color=blue!20!white, caption={}]{\textbf{Adam:} #1}}
\providecommand{\tabularnewline}{\\}

\usepackage{amsthm}
\newtheoremstyle{break}
{}{}
{}{}
{}{}
{ }%
{\thmname{\textbf{#1}}\thmnumber{ \textbf{\@{#2}}}%
  \thmnote{ {\bfseries(#3)}} } 
\makeatother
\theoremstyle{break}
\newtheorem{definition}{Definition}
\usepackage{multirow}
\usepackage{amstext}
\usepackage{stackrel}

\begin{document}

\title{Anticipatory Thinking: A Metacognitive Capability}
 
\author{Adam Amos-Binks}{aamosbinks@ara.com}
\address{Applied Research Associates, Raleigh, NC, 27614 USA}
\author{Dustin Dannenhauer}{dustin.td@gmail.com}
\address{Navatek, Arlington, VA 22203 USA}
\vskip 0.2in
 
\begin{abstract}
Anticipatory thinking is a complex cognitive process for assessing and managing risk in many contexts. Humans use anticipatory thinking to identify potential future issues and proactively take actions to manage their risks. In this paper we define a cognitive systems approach to anticipatory thinking as a metacognitive goal reasoning mechanism. The contributions of this paper include (1) defining anticipatory thinking in the MIDCA cognitive architecture, (2) operationalizing anticipatory thinking as a three step process for managing risk in plans, and (3) a numeric risk assessment calculating an expected cost-benefit ratio for modifying a plan with anticipatory actions.
\end{abstract}

\section{Introduction}
Anticipatory Thinking (AT) is the deliberate and divergent analysis of relevant future states that is a critical skill in medical, military, and intelligence analysis~\citep{Geden2018}. It differs from predicting a single correct outcome in that its goal is to identify key indicators or threatening conditions so one might proactively mitigate and intervene at critical points to avoid catastrophic failure. This uniquely human ability allows us to learn, and act, without actually experiencing. AI systems with this robust capability would support the autonomy and contextual reasoning needed for next generation AI.

However, AI systems have yet to adopt this capability. While agents with a metacognitive architecture can formulate their own goals or adapt their plans in response to their environment (e.g. \cite{Cox2016}) and learning-driven goal generation anticipates new goals from past examples~\citep{Pozanco2018}, they do not reason prospectively about how their current goals could potentially fail or become attainable. Expectations have a similar limitation, they represent an agent's mental view of future states and are useful for diagnosing plan failure and discrepancies in execution~\citep{Munoz2019} but do not critically examine a plan or goal for potential weaknesses or opportunities in advance. To ensure that maintenance goals do not fail, proactive maintenance goals are achieved with plans that do not non-conflict with existing achievement goals, a similar goal to anticipatory thinking, but require an explicit knowledge representation in the domain~\citep{Duff2006} and is more akin to prediction than anticipation. At present, agents do not analyze plans and goals to reveal their unnamed risks (e.g. such as actions of another agent) and how they might be proactively mitigated to avoid execution failures. Calls to the AI community to investigate \textit{imagination machines}~\citep{mahadevan2018imagination} highlights the limitations between current data-driven advances in AI and matching human performance in the long term.

To address this limitation, we take a step towards \textit{imagination machines} with a contribution that operationalizes the concept of anticipatory thinking, a cognitive process reliant on an ample supply of imagination, as a metacognitive capability. We propose this capability as a kind of solution formulation method, a post-planning step that analyzes a solution plan for potential weaknesses and modifies the solution plan to account for them. This approach is in contrast to problem formulation, a pre-planning step that analyzes a problem for efficient search strategies, as well as online risk-aware planning processes \citep{huang2019b}. Our first step of AT identifies properties of a plan that are prone to failure. These include concepts such as atoms needed throughout a plan but are only achieved in the initial state. As a second step, we extend goal-reasoning agent expectations to include anticipatory expectations. A kind of expectation derived from a plan's relevant states that identifies exogenous sources that could potentially introduce failures. Finally, we define anticipatory reasoning to proactively mitigate the potential failures. An agent reasons over the conditions in the anticipatory expectations, generating anticipatory actions to be executed at specific times, foiling an exogenous source of failure.  To exercise this new capability we use a simple example and define metrics for evaluating an agent's anticipatory thinking. 

\section{Previous Work}
Our contributions are based on three related areas of work. Prospective cognition is a fledgling field in cognitive psychology who's goal is to understand human ability to reason about and imagine the future. We discuss some prospection modalities. A second area, goal-reasoning agents, is a type of agent that adapts to and formulates their own goals in response to their environment. We highlight some of the overlap between prospection modalities and the agent's methods for formulating and achieving goals. Finally, investigations in metacognition's role in decision making and behavior has drawn a close tie with autonomy. We detail some of the existing capability to frame anticipatory thinking's role.

\subsection{Anticipatory Thinking}
Anticipatory thinking is an emerging concept in psychology~\cite{Geden2018} that captures the cognitive processes in use when preparing for the future. The deliberate consideration of a diverse set of possible futures differentiates it from a purely imaginative, divergent, or prospective process but rather an aggregate of all three. Imagination is a mechanism to reason about what is outside our immediate sensory inputs. More than an artist's creative reservoir, imagination drives the creativity in complex sciences from engineering to finance. Imagination is used to both reason about details in problem-solving, such as what might have happened in a mystery novel, and as well as generating novel ideas through methods such as counterfactual reasoning. Calls to the AI community to investigate \textit{imagination machines}~\citep{mahadevan2018imagination} highlights the gaps between current data-driven advances in AI and matching human performance in the long term. 

Divergent thinking is often used to assess individual differences in creativity and has been part of scientific studies on creativity since the 1960's~\citep{guilford1967creativity}. Assessing divergent thinking  asks subjects to perform divergent thinking tasks, the scores and measures of which are still the focus of numerous studies~\citep{silvia2008assessing}. Physical limitations such as working memory and recall from long-term memory have been the source of inspiration for developing methodologies to counteract them (e.g. structured analytic techniques~\citep{Heuer2008}.

Lastly, the emerging field of prospection (the ability to reason about what may happen in the future) is driven by future-oriented imagination. Szpunar et al.~\citeyear{Szpunar2014} provide a taxonomy of prospection that covers four modalities (planning, intention, simulation, prediction) in both syntactic and semantic spaces. Several AI research communities have investigated the methods that, at least in name, overlap  with the modalities but have lacked the unifying taxonomy to characterize them in prospective cognition.

\subsection{Goal Reasoning}
One approach to mitigate risks is to encode mappings from states to goals, such that when an agent is in a state, it should pursue the corresponding goal. Thus, if risks are known at design time, an agent can be given mappings from risky states to mitigating goals. MADBot \citep{coddington2005madbot} investigated goal formulation via motivator strategies within, and external to, the planning process. An example of a motivator function is the following: a rover robot may have the motivator function that when it's battery level reaches a threshold, the agent will generate a goal to have a fully charged battery. This would then be achieved by a plan for the rover to navigate to the power source and plug itself in. As shown in Coddington (2005) these motivator functions can be either (1) encoded into the plan operators as constraints (i.e. every action has a precondition that the battery level is above a threshold) or (2) a separate goal formulation process which runs outside the planner and generates a new goals when motivator functions trigger. 

Other approaches to mitigating risk with planning systems include rationale-based monitors \citep{veloso1998rationale}, perceptual-based plan monitors \citep{alavi2018}, and contingency planning \citep{hoffmann2005contingent}. Prior work on mitigating risk during plan execution has considered monitoring rationales for goals \citep{dannenhauer2019anticipation}. In MADBot and other work on goal motivator strategies \citep{munoz2015guiding}, goal motivator functions are known at the design time of the agent. The primary difference of the approach presented here is that goal formulation strategies are identified automatically at runtime by anticipatory thinking approaches using the plan solution as a source of information.

\subsection{Metacognition}
\begin{figure*}[ht]
  \centering
  \includegraphics[width=0.6\textwidth]{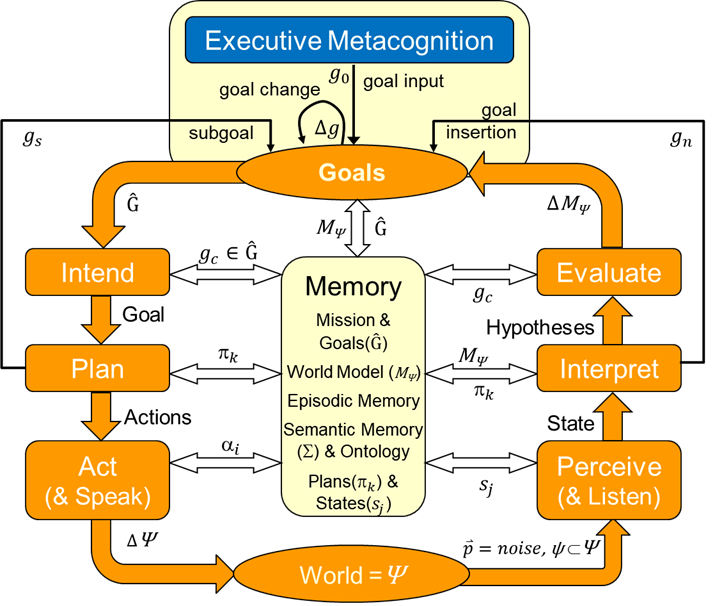}
  \caption{\textit{MIDCA} cognitive level.}
  \label{fig:midca_obj_level}
\end{figure*}

\begin{figure*}[ht]
  \centering
  \includegraphics[width=0.8\textwidth]{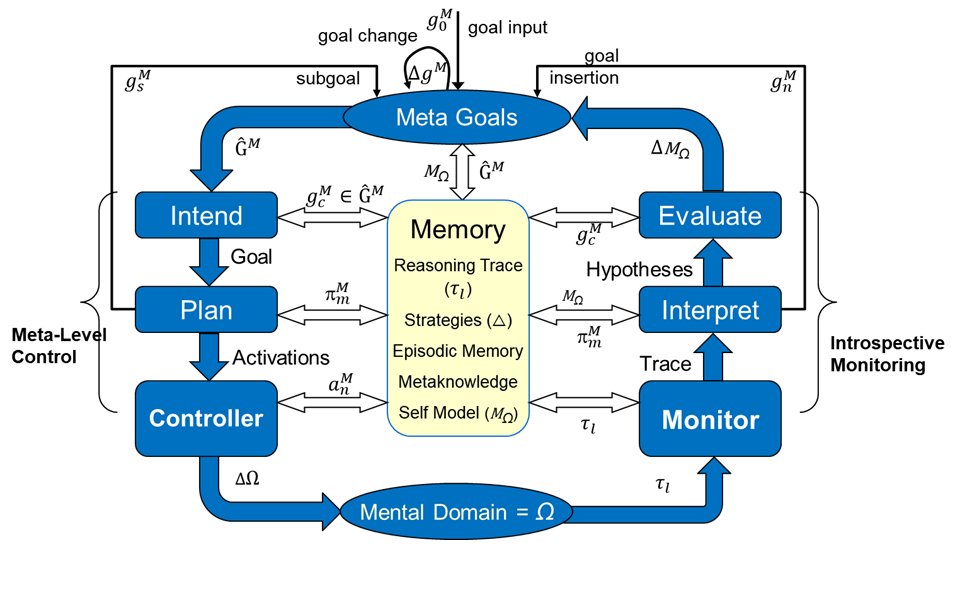}
  \caption{\textit{MIDCA} metacognitive level.}
  \label{fig:midca_meta_level}
\end{figure*}

Metacognition refers to processes that reason about cognition in some form or another \citep{cox2011metareasoning}. We use the Metacognitive Integrated Dual-Cycle Architecture (MIDCA) to discuss anticipatory thinking processes. A primary benefit of MIDCA is it's explicit separation of cognitive and metacognitive processes. Cognitive processes (see Figure \ref{fig:midca_obj_level}) are those that are more directly concerned with the world (goals are world states, plans are sequences of actions that act on world states, etc). Metacognitive processes (see Figure \ref{fig:midca_meta_level}) are those that are more directly concerned with cognitive processes and states (identifying and resolving issues such as impasses that arise in various cognitive processes). One of the core assumptions here is that the agent's mental state is separate from the world state (otherwise reasoning about world states would also be reasoning about cognitive states).

At a general level it seems that AT could be considered a cognitive process since the objective of AT is to prevent risk that arises from various world states in order to achieve some goal that is a world state. When considering specific AT processes (presented in the next section) we argue that AT is truly a metacognitive process since it is concerned with meta goals such as \textit{achieved(\textbf{g'})} where \textbf{g'} is a cognitive level goal. AT is also concerned with decision making on resource trade-offs (a type of metareasoning) for risk mitigation (i.e. spending X extra actions to mitigate Y potential risks). Additionally, if AT processes were to take into account an agent's likelihood of succeeding at a task, than AT processes are making use metacognition self-prediction mechanisms.

MIDCA is currently under active development, and until recently most work has consisted of implementation at the cognitive level. Prior work on the metacognitive level includes monitoring capabilities that maintain a cognitive trace and control actions capable of switching planning algorithms at runtime \citep{cox2017goal} as well as work on domain independent expectations of cognitive processes \citep{Dannenhauer2018}. The primary contributions from these works have mostly focused on the Monitor, Interpret\footnote{The Interpret phase generally includes discrepancy detection, explanation/diagnosis, and goal formulation. Of these, discrepancy detection is the only one with prior work using MIDCA}, and Control phases of the metacognitive layer. The AT process we describe in this paper proposes additional new methods to the Intrepret, Plan, and Control phases of the metacognitive level. 

\section{Anticipatory Thinking as Metacognition}
Our approach to operationalizing anticipatory thinking begins with the concept as explained from~\cite{Geden2018}, "deliberate, divergent exploration and analysis of relevant futures to avoid surprise". We define three steps that operationalize the (i) deliberate, (ii) divergent, and (iii) relevant components of the above AT concept.

First, we identify goal vulnerabilities. This step reasons over a plan's structure to identify properties that would be particularly costly were they not to go according to plan. A second step, failure anticipation, identifies sources of failure for the vulnerabilities. Sources of failure can range from unknown environment states to other agent's interfering goals. Finally, a failure mitigation step modifies an existing plan to reduce the exposure to the sources of failure and creates an anticipatory expectation. We represent these steps in the process in Table~\ref{tab:at_gen} and demonstrate these steps through an \textit{NBeacons} running example from~\cite{Dannenhauer2018}.

\begin{table}[th]
\caption{This method takes as input a plan and returns a modified plan that includes anticipatory actions.}

\noindent \begin{centering}
\begin{tabular}{|c|l|}
\hline 
 & METHOD: Anticipatory Thinking\tabularnewline
 & INPUT: A plan $\pi$\tabularnewline
 & OUTPUT: A modified plan $\pi^{\prime}$ \tabularnewline
\hline 
1 & identify goal vulnerabilities in $\pi$\tabularnewline
2 & identify failure anticipation in $\pi$\tabularnewline
3 & edit $\pi$ with failure mitigations to get $\pi^{\prime}$ \tabularnewline
4 & return $\pi^{\prime}$ \tabularnewline
\hline 
\end{tabular}
\par\end{centering}
\label{tab:at_gen}
\end{table}

In the NBeacons domain, an agent must generate plans to reach beacons and activate them. If the agent ever passes through a sandpit square, they must take three actions to dig out. The wind may blow in a known direction at a known speed after every agent action. The wind pushes an agent a number of squares further (equivalent to the speed) in the wind's direction and can result in an agent passing over a sandpit and getting stuck. In our example, the wind is blowing West at a speed of five making an agent's plan vulnerable to any sandpit that lies within five squares West of their location.

We use the Partial Order Causal Link (POCL) representation~\citep{Penberthy1992} for an agent's plans. The main advantage to using POCL over other plan representations is that \textit{causal link threats} can explicitly represents potential failures from external events. We use the typical definitions for POCL representation from \cite{Penberthy1992} where a POCL plan consists of steps (S) that are ground actions from the domain model, bindings (B) that map free variables to literals, step orderings (O) that constrain when steps must execute relative to one another, and causal links (L) connect steps to one another when an effect of one step instantiates a precondition for a following step. While the above plain english definitions of POCL representation will suffice for those who are familiar with planning, we provide a formal definitions for causal link threats as they are key to our choice of POCL.

\begin{definition}[Causal link threat] A causal link threat occurs when a causal link is established $s\stackrel{p}{\rightarrow}u$, and some other step $w$ has effect $\neg p$ and could be executed after $s$ but before $u$. Executing $w$ in this interval means the precondition $p$ of $u$ is no longer satisfied by the state after $s$ is executed and thus $u$ will not execute. \label{def:causal_link_threat}
\end{definition}

\subsection{Goal Vulnerabilities}\label{sec:at_step_1}
Identifying a plan's vulnerable structural properties is the first step in proactively mitigating its failure. We define a single vulnerability for our \textit{NBeacons} example, of what could be numerous vulnerable properties of a plan. 

Our first property, precondition strength, is a measure of how many times a precondition is established and used in plan. The fewer times a precondition is established and the more it is used increases the vulnerability of a plan to failure.


\begin{definition}[Precondition Strength] 
Precondition strength of a plan, \textsc{PreStrength}($\pi$), is a set of tuples $\langle a,p,e \rangle$ where $a$ is a literal, $p$ the number of steps in $\pi$ that use it as a precondition, $e$ the number of times it is an effect before first used as precondition. \label{def:pre_strength} 
\end{definition}

\begin{figure*}[ht]
  \centering
  \includegraphics[width=0.5\textwidth]{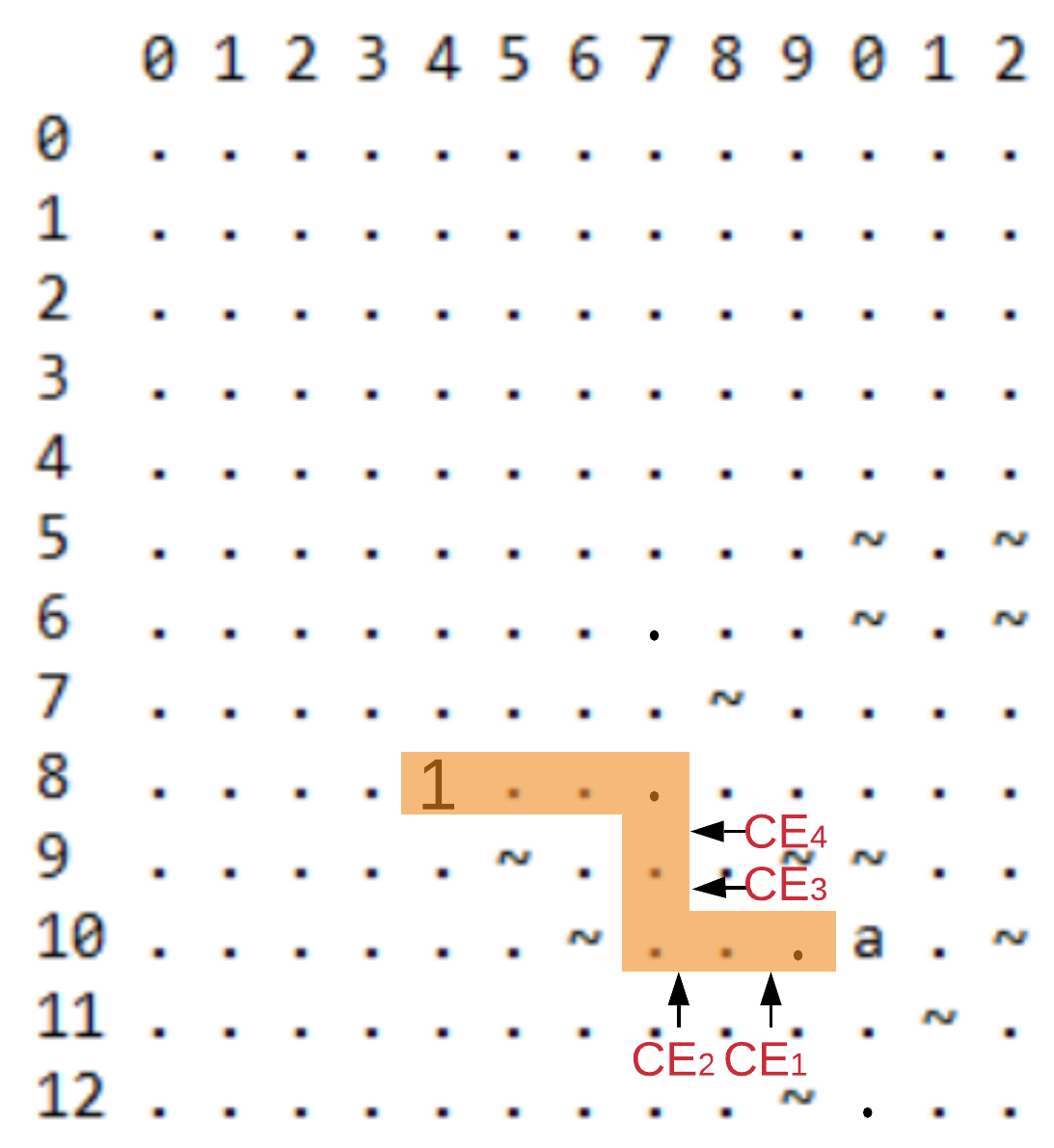}
  \caption{An \textit{NBeacons} example that shows an optimal path in orange along with four conditioning events that could take place.}
  \label{fig:nbeacons_grid}
\end{figure*}

Our agent's existing plan (the orange path in Figure~\ref{fig:nbeacons_grid} comprises of eight move actions all of which have the (canMove) precondition. This precondition is only established once in the initial state and so its entry in \textsc{PreStrength}($\pi$) is $\langle canMove(agent), 8, 1 \rangle$.

\subsection{Failure Anticipation}\label{sec:failure_anticipation}
Vulnerabilities are not by themselves indicative that a plan is at risk of failure. Risk of failure requires some means to exploit the vulnerability, what we will call conditioning events. We approach identifying these events as a kind of prefactual reasoning (future-oriented counterfactual reasoning), where we take the negation of the most vulnerable preconditions (identified in Section \ref{sec:at_step_1}) and identify actions with them as effects.


\begin{definition}[Conditioning Event]
The conditioning events of a plan, $\textsc{CE}(\pi)$ are actions in the domain model such that one or more effects of each action is the negation of a precondition of a step in $\pi$, introducing a causal link threat. \label{def:cond_event}
\end{definition}

In our \textit{NBeacons} example, wind serves as a conditioning event. If the wind blows at any point a sandpit is five squares West (or less) from the agent, a causal link threat is introduced as \textit{canMove(agent)} will not be true and the agent will be forced to spend three actions digging out of the sand pit. This results in four potential conditioning events, one after each of the agent's first four moves, indicated by the red CE notation in Figure~\ref{fig:nbeacons_grid}.

\subsection{Failure Mitigation}\label{sec:failure_mitigation}
The final step to operationalizing anticipatory thinking is to define what the relevant property means for a goal-reasoning agent. We term this step failure mitigation where the agent reasons over conditioning events to identify actions that reduce a plan's risk exposure to these conditioning events. These action are anticipatory actions.

\begin{definition}[Anticipatory Actions]
The anticipatory actions of a plan, $\textsc{ANT}$, is a set of tuples $\langle A^{\textsc{ANT}}, \textsc{CE} \rangle$ where $A^{\textsc{ANT}}$ is an action sequence, $a_i, a_2, ...a_n$ added to $\pi$ such that at least one effect reduces the impact of the conditioning events $\textsc{CE}$ in $\textsc{CE}(\pi)$. \label{def:ant_action}
\end{definition}

\begin{figure*}[ht]
  \centering
  \includegraphics[width=1.0\textwidth]{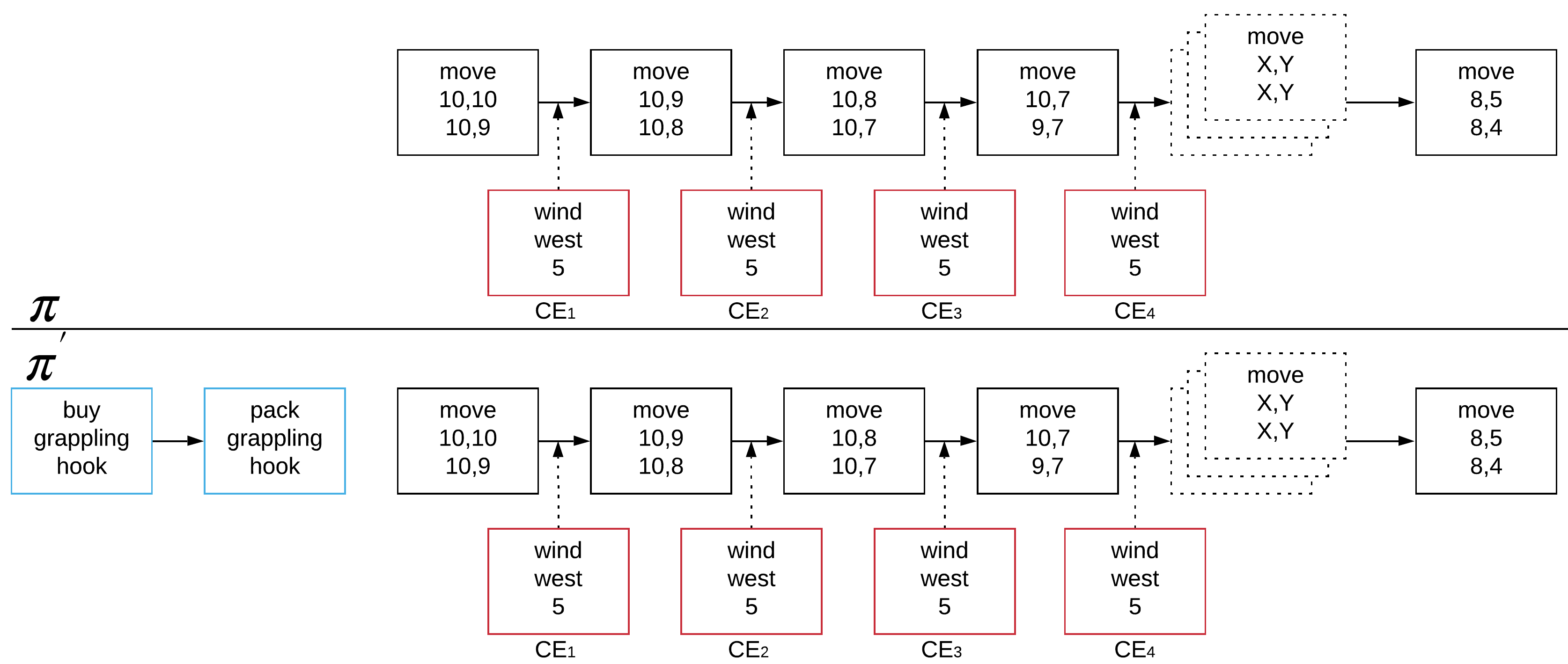}
  \caption{The original plan ($\pi$) on the top with conditioning events and modified plan ($\pi^{\prime}$) with two anticipatory actions (blue) on the bottom }
  \label{fig:plans}
\end{figure*}

To mitigate the unpleasantness of being blown about by the wind and getting trapped in a sandpit, our agent has the option to outfit itself with a grappling hook. A grappling hook allows an agent to move out of a sandpit in a single action. However, adding the grappling hook adds action costs of one for the \textit{buy} and \textit{pack} steps that need to be executed. We add these two anticipatory actions to the agent's plan before the journey begins, see $\pi^{\prime}$ in Figure~\ref{fig:plans}.

Adding the grappling hook to the plan creates an expectation within an agent that risk exposure to the wind conditioning event has been reduced. We refer to this new type of expectation as an anticipatory expectation and define it as:


\begin{definition}[Anticipatory Expectations]
An Anticipatory Expectation is the action cost reductions expected from introducing anticipatory actions to mitigate conditioning events. \label{def:ant_expectation}
\end{definition}


\subsection{Anticipatory Thinking in MIDCA}\label{sec:at_in_midca}
We now put forth an anticipatory thinking approach as a metacognitive process in MIDCA, highlighting the role of each phase of the metacognitive layer:
\begin{description}
\item[Monitor:] Obtain observations of the cognitive level components, including the current plan $p$ and the current goal $g$.
\item[Interpret (as composed of the following three steps):]
~

\begin{description}
\item[Discrepancy Detection:] Flag the current plan $p$ from the cognitive level Plan phase (see Figure 1, cognitive layer, left side) as potentially risky, \textit{risk\_level($p$, HIGH)}. 
\item[Explanation / Diagnosis:] Assess the risks associated with the plan $p$ using anticipatory thinking approaches, such as those described in Section \ref{sec:at_step_1} using the notions of \textit{prestrength}. The results of the analysis would be vulnerabilities $V$ of the plan $p$.
\item[Goal Formulation:] Formulate the goal to transform $p$ into a new plan $p'$ with a safer risk level, while maintaining that the current goal of $p$ is achieved. The new goal would then be \{\textit{risk\_level($p'$, LOW)} $\land$ $achieves(p',g) $\}.
\end{description}
\item[Evaluate:] Drop any meta goals if they have been achieved.
\item[Intend:] Commit to achieving the newly formulated meta goal \{\textit{risk\_level($p'$, LOW)} $\land$ $achieves(p',g) $\}.
\item[Plan:] Take current mental state containing \textit{risk\_level($p$,HIGH)} and vulnerabilities $V$ and search for a set of new actions, meta\_plan $m_p$, consisting of add or delete edits from plan $p$ in order to achieve $p'$ such that \{\textit{risk\_level($p'$, LOW)} $\land$ $achieves(p',g) $\}. 
\item[Control:] Carry out the sequence of plan edits in $m_p$ resulting in a new $p'$ such that \textit{risk\_level($p'$,LOW)} and $p' \models g$ where $g$ is the original goal of $p$.  
\end{description}

The primary effort occurs in the Plan phase which we speculate could be modeled as a search process such that nodes are plans and their associated risk\_levels and edges between nodes are anticipatory actions that are added to (or possibly removed from) the plan. The search process would terminate when a goal node is reached that meets a low risk\_level for the plan inside the node. This example through the metacognitive phases serves as one possible realization of AT in MIDCA. We leave more concrete implementation details for future work.

\section{Evaluation Framework}
Anticipatory thinking is concerned with identifying possible worlds that affect desirable outcomes and taking action to mitigate them. This differs from typical future-oriented analysis centered around prediction that are focused on identifying a single likely outcome. As such, to appropriately evaluate anticipatory thinking we require alternative measures than those used in prediction.

\subsection{Successful Anticipatory Thinking}
Conceptually, anticipatory thinking's goal is to have a high recall rate. That is, to ensure the future that unfolds is accounted for in a set of possible futures. However, calculating recall does not capture the cost of adding anticipatory actions or the cost of identifying conditioning events. To address this limitation we develop an assessment of anticipatory thinking that accounts for the cost in relation to the potential benefits. 

\begin{figure*}[ht]
  \centering
  \includegraphics[width=0.5\textwidth]{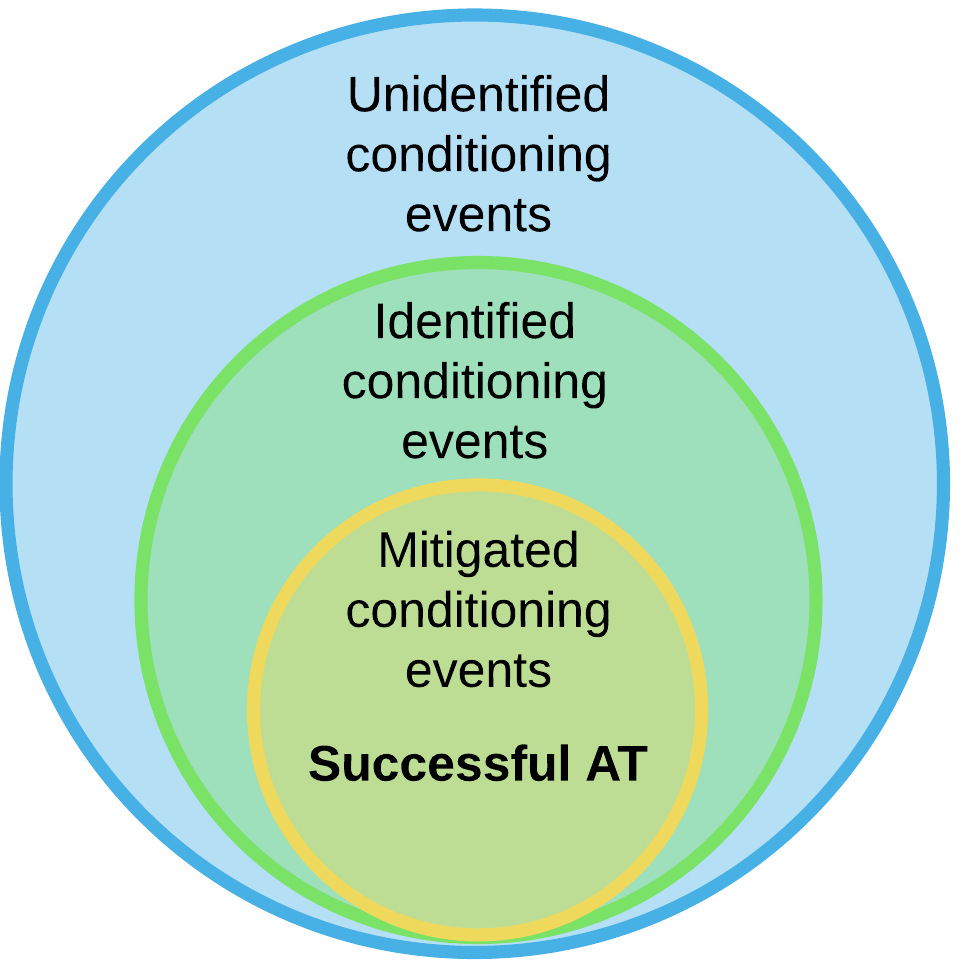}
  \caption{Successful anticipatory thinking identifies conditioning events that can be mitigated with anticipatory actions. In our \textit{NBeacons} example, before the agent has done any anticipatory thinking, the \textit{wind} conditioning events are \textit{unidentified} and are in the blue area. After the the goal vulnerability and failure anticipation steps (Sections ~\ref{sec:at_step_1},~\ref{sec:failure_anticipation}), \textit{wind} conditioning events are now in the green area. Once anticipatory actions have been added to the plan (Section~\ref{sec:failure_mitigation}), the \textit{wind} conditioning events are mitigated and in the yellow area.}
  \label{fig:at_venn}
\end{figure*}

An additional challenge is to avoid coupling anticipated outcomes to the actual outcomes. AT mitigates, not predicts, failure. Therefore AT assessment should only assess the potential payoff from mitigating, not whether any individual future comes to pass.



In Figure~\ref{fig:at_venn}, we represent anticipatory thinking as a plan's identified conditioning events in the green circle. Before anticipatory thinking, conditioning events are unknown to our agent and reside in the blue area. Successful anticipatory thinking is the set of identified conditioning events where anticipatory actions are taken to mitigate their impact and reside in the yellow area. We assess successful anticipatory thinking as 
\begin{equation}
AT^{assess}(\pi)=\frac{|CE(ANT)|}{|CE|}\times\left(1-\frac{|A^{\textsc{ANT}}(\textsc{ANT}_{i})|}{\stackrel[i]{|\textsc{ANT}|}{\sum}\left(\stackrel[j]{|CE(\textsc{ANT}_{i})|}{\sum}impact(ce_{j})\right)}\right),
\label{eq:at_assess}
\end{equation}
where $|CE|$ is the number of conditioning events identified and $|CE(ANT)|$ are the mitigated conditioning events. Their ratio represents how many conditioning events were mitigated. A second ratio calculates the potential benefit of mitigation. For each anticipatory action set, $\textsc{ANT}$, we sum the impact of each conditioning event mitigated, $impact(ce_{j})$, and subtract the cost of the anticipatory actions, $AA(\textsc{ANT}_{i})$.

\subsection{Example}
Applying equation~\ref{eq:at_assess} to our NBeacons conditioning events, we have four wind conditioning events, $|CE|=4$, and each one is mitigated, $|CE(ANT)|=4$. This ratio of 1.0 ($4/4$) is best possible case in that every identified conditioning event was mitigated. Conversely, plans where many identified conditioning events that have few mitigations would have a ratio closer to zero and may benefit from the use of robust search algorithms. Our next ratio assesses the potential mitigation benefit. Mitigating the wind event requires buying and packing the grappling hook, each with an action cost of one for a total of two, $|AA(\textsc{ANT}_{i})|=2$. The sole anticipatory action sequence, $i=1$, is expected to save the agent three dig actions for each of the four wind events, $j=4$, resulting in a potential mitigation of twelve actions, and a resulting mitigation ratio of 0.83 ($1-0.17$). Again, plans with not so favorable benefits from mitigations would have a lower expected payoff from their actions and would have a ratio closer to zero. Together these two ratios result in an $AT^{assess}(\pi^{\prime})$ of 0.83 ($1.0\times-0.83$), this calculation is reflected in Equation~\ref{eq:at_assess_pi}.

\begin{equation}
AT^{assess}(\pi^{\prime})=\frac{4}{4}\times\left(1-\frac{2}{\stackrel[i]{1}{\sum}\left(3+3+3+3\right)}\right),
\label{eq:at_assess_pi}
\end{equation}

\section{Conclusion and Future Work}
Anticipatory thinking is a complex cognitive process for assessing and managing risk in many contexts. It allows humans to identify potential future issues and proactively take actions in the present that will manage their risks. We have defined how an artificial agent may perform anticipatory thinking at a goal reasoning level, so they may receive the same benefits and enable further autonomous capability.

Our approach made three contributions. First we defined anticipatory thinking in the MIDCA cognitive architecture as a goal reasoning process at the metacognitive layer. Specifically, Section \ref{sec:at_in_midca} highlights the role of AT in each phase of the metacognitive layer of MIDCA shown in Figure \ref{fig:midca_meta_level}. Second, we operationalized the anticipatory thinking concept as a three step process for managing risk in plans. Goal vulnerabilities, failure anticipations and failure mitigation identify weakness of a plan, their potential failure sources (conditioning events), and failure mitigations (anticipatory actions) to reduce the impact of the failure sources. Finally, we proposed a numeric assessment for successful anticipatory thinking. Key to the assessment are a ratio of identified conditioning events to mitigated ones and an expected cost-benefit ratio for the anticipatory actions.

We expect two immediate areas of future work. First, we are planning to integrate our anticipatory thinking definitions into an existing MIDCA implementation. From there, we will be able to perform experiments on existing domains. A second area is to develop more methodologies for each of the three anticipatory thinking steps. Expanding the failure sources beyond the failure inducing step (e.g. an action sequence) to identify the most parsimonious mitigation and extracting some benefit from unmitigated conditioning events are promising avenues of investigation.


\vspace{-0.25in}

{\parindent -10pt\leftskip 10pt\noindent
\bibliographystyle{cogsysapa}
\bibliography{format}

}


\end{document}